\pdfoutput=1
\documentclass[11pt]{article}
\usepackage[table]{xcolor}
\usepackage[]{EMNLP2022}
\usepackage{graphicx,subfigure}
\usepackage{times}
\usepackage{latexsym}
\usepackage{comment}
\usepackage{wrapfig}
\usepackage{float}
\usepackage{amsmath}
\usepackage{multirow}
\usepackage{float} 
\usepackage{booktabs}
\usepackage[linesnumbered,ruled,vlined]{algorithm2e}
\usepackage{algpseudocode}
\usepackage{adjustbox}
\usepackage{comment}
\usepackage{MnSymbol}
\usepackage[utf8]{inputenc}

\usepackage{microtype}

\usepackage{inconsolata}

\title{\includegraphics[width=0.3in]{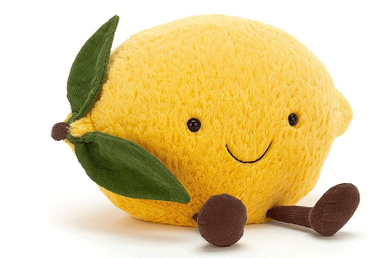} LEMON: LanguagE ModeL for Negative \\ Sampling of Knowledge Graph Embeddings}

\author{
Md Rashad Al Hasan Rony$^*\filleddiamond$ \\
  University of Bonn \\
  BMW Group \\
  \texttt{md-rashad-al-hasan.rony@bmwgroup.com} \\\And
 \rule{0.95in}{-6pt} Mirza Mohtashim Alam$^*$\\
  \rule{0.95in}{-6pt}Nature-Inspired Machine Intelligence-InfAI\\
  \rule{0.95in}{-6pt}\texttt{mohtasim@infai.org} \\\And
\rule{1.4in}{-6pt}Semab Ali \\
\rule{1.4in}{-6pt}University of Bonn\\
\rule{1.4in}{-6pt}\texttt{s6sealii@uni-bonn.de} \\\AND
  \rule{-0.6in}{-6pt}Jens Lehmann \\
  \rule{-0.6in}{-6pt}Nature-Inspired Machine Intelligence-InfAI\\
  \rule{-0.6in}{-6pt}\texttt{lehmann@infai.org} \\\And
  \rule{0.6in}{-6pt}Sahar Vahdati\\
  \rule{0.6in}{-6pt}Nature-Inspired Machine Intelligence-InfAI\\
  \rule{0.6in}{-6pt}\texttt{vahdati@infai.org}}
\begin{document}

\maketitle
\def\thefootnote{*}\footnotetext{These authors contributed equally to this work\\ $\filleddiamond$ work was done when the author was at Fraunhofer IAIS}\def\thefootnote{\arabic{footnote}}
\begin{abstract}
Knowledge Graph Embedding models have become an important area of machine learning. 
Those models provide a latent representation of entities and relations in a knowledge graph which can then be used in downstream machine learning tasks such as link prediction.
The learning process of such models can be performed by contrasting positive and negative triples.
While all triples of a KG are considered positive, negative triples are usually not readily available. 
Therefore, the choice of the sampling method to obtain the negative triples play a crucial role in the performance and effectiveness of Knowledge Graph Embedding models. 
Most of the current methods fetch negative samples from a random distribution of entities in the underlying Knowledge Graph which also often includes meaningless triples. 
Other known methods use adversarial techniques or generative neural networks which consequently reduce the efficiency of the process. 
In this paper, we propose an approach for generating informative negative samples considering available complementary knowledge about entities. 
Particularly, Pre-trained Language Models are used to form neighborhood clusters by utilizing the distances between entities to obtain representations of symbolic entities via their textual information. 
Our comprehensive evaluations demonstrate the effectiveness of the proposed approach on benchmark Knowledge Graphs with textual information for the link prediction task.
\end{abstract}
\section{Introduction}

Knowledge graphs (KG) are a data model to represent the facts from the real world in the form of triple $(h,r,t)$ where $h$ and $t$ refer to head and tail entities, and $r$ denotes the relations between them. 
Despite the size of some KGs reaching billions of triples, they usually remain incomplete.
This is due to the difficulties in capturing all relevant facts about a domain of interest at the time of KG construction.
% capturing all relevant facts of a focused domain of interest.
%Therefore, the last decade witnessed various AI-based approaches for KG completion. 
Knowledge Graph Embedding (KGE) models are the most prominent approach to complete KGs.
%In recent years, graph representation learning has gained significant research attention due to the growing number of knowledge graph based applications such as link prediction~\cite{alameswc,threed} and node classification~\cite{}. 
The learning process of KGEs involves contrasting positive and negative triples which plays an important role for the effectiveness of the models as well as the ultimate results on downstream tasks. 
%However, KGs follow closed-world assumption meaning that ``all the present triples are considered to be positive (correct)''.
%In most of KGE models, negative samples have been generated from the existing positive triples through a random triple corruption method.
%In this way, the significant role of negative sampling and its impact on the overall performance are not carefully considered.
%This is majorly due to the restrictions of the random distribution that affects the model generalization, specially on unseen data.
%Additionally, negative sampling is considered to be an efficient and adaptive technique for training KGEs, since it does not treat all the unobserved data as negative.
%
%\vspace{-0.5mm}
%
Despite the importance of the methods for sampling negative triples from the existing positive triples in a KG, only recently this has gained attention and some techniques have been proposed to increase their efficiency~\cite{mikolov2013distributed,sun2018rotate,cai-wang-2018-kbgan,wang2018incorporating,dash2019distributional,Alam2020AffinityDN}.
However, still most of these methods only fetch the negative samples (NS) through a random distribution including uniform distribution~\cite{he2017neural} or population-based distribution~\cite{chen2017sampling} from the entities in the underlying KG.
The drawback of these approaches is that the negative samples also include a high percentage of meaningless triples, i.e., the classification problem of distinguishing positive and negative triples becomes (too) simple. 

% \begin{figure}
% %\centering
%   \includegraphics[scale=0.11]{images/example3.png}
%   \caption{Lemon vs. Random Distribution}
%   \label{fig:exampel}
% %  \centering
% \end{figure}

%Other known methods use adversarial techniques or generative networks which add complexity burden on the models. 
%The initial method named Uniform~\cite{he2017neural} proposed a fixed distribution and another one uses a population-based distribution~\cite{chen2017sampling}. 
In other works, Generative Adversarial networks (GANs)~\cite{goodfellow2014generative} are used for generating negative samples for KGEs. 
Most of these works are computationally expensive as they either suffer from the vanishing gradient problem or require a high number of training parameters.
Furthermore, these approaches focus on learning the graph patterns rather than considering any complementary knowledge that is carried by entities and relations. 
% In Figure~\ref{fig:exampel}, a candidate triple $(molecular\_function, produces, vitamin)$ is used for generating negative samples by corrupting the tail entity. 
% A uniform distribution could, for example, replace $vitamin$ by $action$ in order to generate a negative triple $(molecular\_function, produces, action)$. 
%and $(molecular\_function, produces, Hormones)$, respectively (lower part of Figure \ref{fig:exampel}). 
% As visible, the first triples is far from being effective negative samples as the generated triple is meaningless.
%This triple, however, is meaningless as the tail should rather be a molecule or at least any type of substance. 
Generating such negative samples can reduce the performance of KGE models and affect their link prediction capability. 
Especially in large KGs, it is very likely that most negative samples are sampled from a very large distribution. 

\begin{figure*}
\centering
  \includegraphics[scale=0.3]{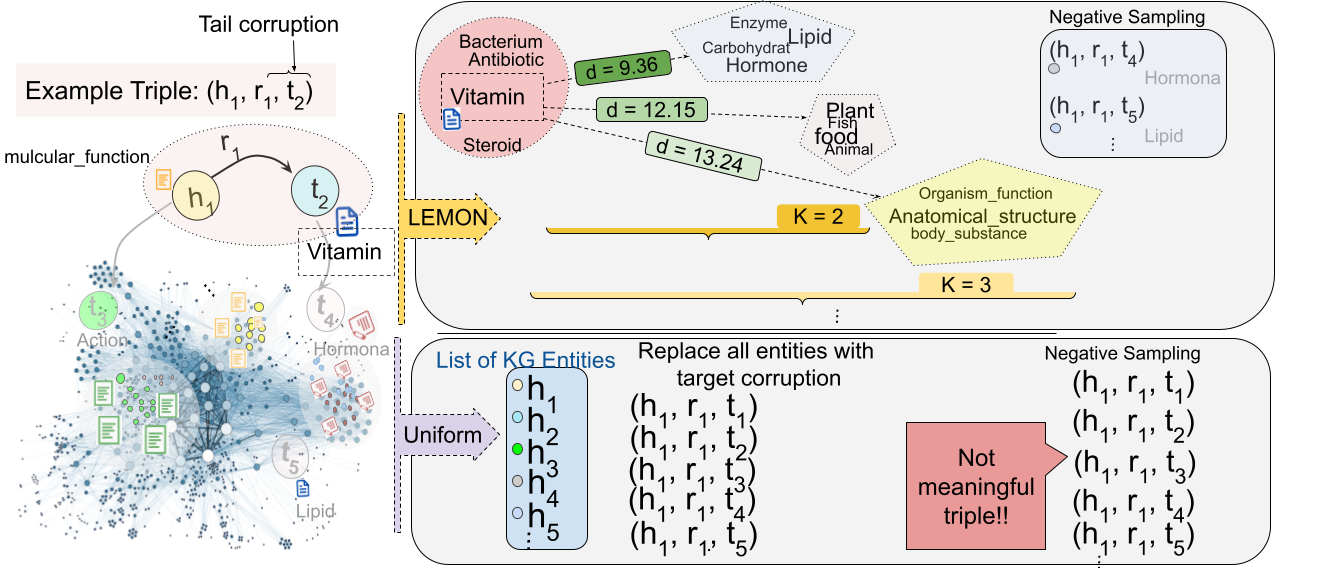}
   \caption{Lemon vs. Random Distribution}
   \label{fig:exampel}
%  \centering
\end{figure*}
To alleviate the aforementioned issues, we propose LEMON, a novel method for negative sample generating in KGE models using Pre-trained Language Models (PLMs).
%for the purpose of generating negative samples in KGE models.
The core idea is that the vector representations of contextual knowledge for each entity are obtained from PLMs and influence the similarity of entities.
%and play a significant role in defining the similarities of the entities. 
% Additionally, we leverage K-means++ in a clustering task where the contextualized embedding of KG entities are transferred into a dense or vectorized space to generate negative samples.
%Besides providing semantically similar negative samples, LEMON is also runtime efficient. 
% Besides being runtime and performance efficient, LEMON also provides semantically similar negative samples.
%The motivation for obtaining the negative samples based on the output of PLMs highly relates to the richness of contextual knowledge associated to entities and relations. 
%While, the KGE models learn the vector representation of the entities considering the structure of KG -- ignoring complementary knowledge (in most of the cases), PLMs are not designed to consider such structural knowledge.
%Unfortunately, contextual meanings are often overlooked. 
%In recent years, Transformer~\cite{vaswani2017attention}-based language models have revolutionized the field of NLP. 
%A pre-trained language model is rich in contextual information, and better captures the meaning of a text. 
% Therefore, PLMs are often utilized to tackle various downstream tasks such as dialogue systems~\cite{chaudhuri2021grounding} and natural language inference~\cite{2020t5}. 
%A pre-trained language model,
The choice of PLMs for this work is \textit{Sentence-BERT}~\cite{reimers-2019-sentence-bert} as pre-trained language model to obtain the contextual representation of KG entities. Sentence-BERT is chosen due to its ability to derive semantically meaningful sentences with high computational efficiency. 
% Sentence-BERT is found to be 
%effective for capturing the contextual meaning, because of its Transformer~\cite{vaswani2017attention} architecture.
% Several research works~\cite{rony2022rome,d2020bert} reported the superior performance of Sentence-BERT over other pre-trained word vectors (i.e., Fasttext~\cite{bojanowski2017enriching}, Word2Vec~\cite{mikolov2013distributed}). 
% Furthermore, we apply a dimensionality reduction algorithm to transform the high-dimensional embedding into low-dimensional representation. 
% The low-dimensional contextualized entity embeddings are then clustered utilizing the K-means++ algorithm.
Furthermore, a dimensionality reduction algorithm is used together with the K-means++ clustering technique in order to measure the similarity of entities in a run-time efficient manner. 
%closer together to each other into a dense space in a run-time efficient manner. %This facilitates a run-time efficient negative sampling. 
%Since contextualized meaning is present in the PLMs already, less number of negative sample is required in the training phase.
The embeddings obtained from the language models convey contextual meaning as such models are already trained on large corpora. 
%We argue that such meaning-based negative samples lead to better representation learning. %, confirmed by the evaluation results. 
Specifically, for an entity (either as the head or tail of a triple) to be corrupted, the model considers the  corresponding textual information and identifies an ideal set of clusters to which the target entity belongs. 
When considering the nearest clusters, number of hops are the set of clusters (containing entities) in the desired ranges. From These clusters the negative candidate entities are sampled.
%(see upper part of Figure \ref{fig:exampel}). 
%The identified cluster is designated as the source cluster  %(or a desired nearest clusters)
%to sample negatives entities.
%other than positive entity to be corrupted. 
 
Figure~\ref{fig:exampel} demonstrates the difference between LEMON and a random negative sampling approach. 
In order to corrupt a particular tail entity (in this case $Vitamin$), the distance between the candidate entity and the clusters (containing entities) is crucial. 
The upper part of the figure demonstrates the neighborhood between the clusters based on the various distances $d=\{9.36, 12.15, 13.24\}$ from the candidate entity $Vitamin$. 
Each value of $d$ represents the distance between candidate entity (which is to be corrupted) and K-th cluster. 
The model chooses contextually meaningful entities from the clusters using the desired distances. 
The lower part of the figure illustrates the sampling method of random corruption and the possibility of sampling meaningful entities.
%It is noteworthy that the distance to the other clusters from the source cluster plays a vital role in the training process. % which is also studied

The key contributions of this paper can be summarized as follows: 
\begin{itemize}
  %\item To the best of our knowledge, we are the first to employ and investigate pre-trained language models for negative sampling approach.
  \item A novel approach named LEMON is proposed that employs pre-trained language models and clustering methods to obtain meaningful entities to sample negative triples. 
  To the best of our knowledge, we are the first to employ and investigate pre-trained language models for the purpose of negative sampling approach in KGEs.
  %\item We conduct experiments to evaluate the performance of our proposed method. 
  %We evaluated
  \item A comprehensive evaluation across standard publicly available benchmarks namely WN18, and WN18RR that have sufficient textual information have been performed. 
  %\item Since our negative sampling method is independent of KGE models,
  The evaluation contains several widely used KGE models including TransE~\cite{bordes2013translating}, RotatE~\cite{sun2018rotate}, DistMult~\cite{yang2014embedding}, %ComplEx~\cite{trouillon2016complex} and %QuatE~\cite{zhang2019quaternion}.  Our proposed method achieves state-of-the-art performance on the benchmark datasets.
 %\item We provide analysis that demonstrates the effectiveness of PLMs in combination with KGE models. 
\end{itemize}

%Since our approach is model independent, we experimented the proposed negative sampling method on widely used KGEs in the case of TransE~\cite{bordes2013translating}, and RotatE~\cite{sun2018rotate}. 
%and DistMulti~\cite{yang2014embedding}.
% An extensive evaluation demonstrates the effectiveness and efficiency of our proposed negative sampling method.

\section{Related Work}
Here, we provide a summary of relevant approaches which we classified as following:

\textbf{Distribution-based Approaches.} Many of the initial negative sampling approaches follow random distribution for selection of entities to be corrupted~\cite{mikolov2013distributed,sun2018rotate}. 
%from a pool of candidates
%triples
%Later, an improved sampling technique based on 
Besides uniform~\cite{he2017neural} and population-based distributions~\cite{chen2017sampling}, Bernoulli distribution is considered to limit the appearance of false negative triples in the existing relations~\cite{wang2014knowledge}.
Despite their simplicity, such approaches suffer from the vanishing gradient problem as described in~\cite{cai-wang-2018-kbgan,wang2018incorporating}, and the provided negative samples are not informative.

\textbf{Generative Adversarial Network-based Approaches.} 
Recently, Generative Adversarial Networks (GANs)~\cite{goodfellow2014generative} have been explored for negative sampling to overcome the limitations of fixed distribution based sampling~\cite{cai-wang-2018-kbgan,wang2018incorporating}. 
The discriminator is trained to minimize the margin-based ranking loss, while the generator learns to sample high-quality negative samples~\cite{cai-wang-2018-kbgan}. 
Although such techniques are capable of generating high-quality negative samples, they are expensive to train as addressed in~\cite{zhang2019nscaching}.
%e is required for these methods to learn the full distribution of negative triples. 
%In \cite{zhang2019nscaching}, this issue is addressed by proposing a distilled version to reduce the number of parameters and training time. 
It uses a cashing mechanism which solved the time-efficiency issues, but does not improve the effectiveness of the generated NS. 
% It stores the high scoring negative triples in a cache to get rid of the highly skewed negative sampling distribution issue that exists in the datasets. 
% The pre-processed cache is later used to sample negative triples with high score to tackle the vanishing gradient problem and high training time problem.

\textbf{Structure-aware Approaches.} More recently, a structure-aware negative sampling technique (SANS) was proposed~\cite{ahrabian-etal-2020-structure} to consider the neighbourhood information for generating negative triples. 
To corrupt an entity of a triple for negative sampling, all the k-hop neighborhood nodes connected to the entity are considered negative. Thus, the negative triples are generated from the structural information (i.e., the k-hop neighborhood). 
However, it is crucial for the KGE models to know about possible connections beyond k-hop to perform link prediction. 

\textbf{Other Approaches.}
Data augmentation methods are recently used in negative sampling~\cite{huang2021mixgcf}. 
Graph neural network-based positive mix and hop-mix strategies have been proposed to generate negative samples for a set of neighborhood hops. 
Following a similar research line, MixKG~\cite{che2022mixkg} introduced an approach for generating hard negative samples with a similarity based negative triples. 
These methods are not structure-based nor PLM-based. 
However, the performance of such methods decreases with the increased size of hard negative samples which limits the scaling capability of the model.

% In contrast to the previous works, we employ contextualized representation of the entity and relation and utilized a K-means++ algorithm to cluster similar entities and relations into a dense space. The contextualized vector representation and the clustering method allow our system to effectively and efficiently generated negative samples.
\section{Approach: LEMON}
\label{sec:approach}
% Generation of negative sampling plays a significant role for the performance of KGE models. The impact on the performance is obvious when the obtained negatives are fetched in a systematic way. 
% Many systematic negative samples require a huge computational burden either on the preprocessing or in the negative sampling generation side. Emphasising on the computational burden, our model is more simpler and yet efficient.   

% \begin{figure*}
% \centering
%   \includegraphics[width=0.75\textwidth]{System Diagram of KR paper.png}
%   \caption{Overall Approach}
%  \centering
% \end{figure*}

%In this section, we propose a novel method of negative sampling generation that uses language model and neighborhood clustering for negative sample generation.
In this section, we describe the details about LEMON on how it leverages a pre-trained language model and neighborhood clustering algorithm to generate negative samples for training KGE models.
As the initial step, the embedding of the entity label (text) are obtained from Sentence-BERT~\cite{reimers-2019-sentence-bert}. %Obtaining the approximation of optimal number of cluster is optional in our case since the Elbow method~\cite{Syakur_2018} considered as a hyperparameter.
Furthermore, to cluster similar entities together for building a neighborhood, the cluster size needs to be defined which is considered as one of the key hyperparameters for obtaining the clusters. %We determine the optimal cluster size following the Elbow method~\cite{Syakur_2018}.
Moreover, Principal Component Analysis (PCA)~\cite{abdi2010principal} dimensionality reduction techniques%, Spectral Embeddding~\cite{Luxburg07atutorial} and TSNE~\cite{van2008visualizing} are 
is employed to adjust the dimension of the PLM embeddings. %from Sentence Transformer. 
% Empirically, we achieved better model and runtime performance with lower dimensional embeddings in our setting. 
This is done to eradicate the curse of dimensionality~\cite{verleysen2005curse}, which is often caused by large embedding dimensions (in our case Sentence-BERT embeddings). 
Finally, the K-means++~\cite{arthur2006k} algorithm is leveraged to build the neighborhood clusters. Our proposed method consists of two phases: Building the neighborhood clusters, and negative sample generation. The phases are described in details below.

\subsection{Building the Neighborhood Clusters}
\label{sub: building the neighborhood}
\begin{figure*}[]
    \centering
    \includegraphics[width=0.9\textwidth]{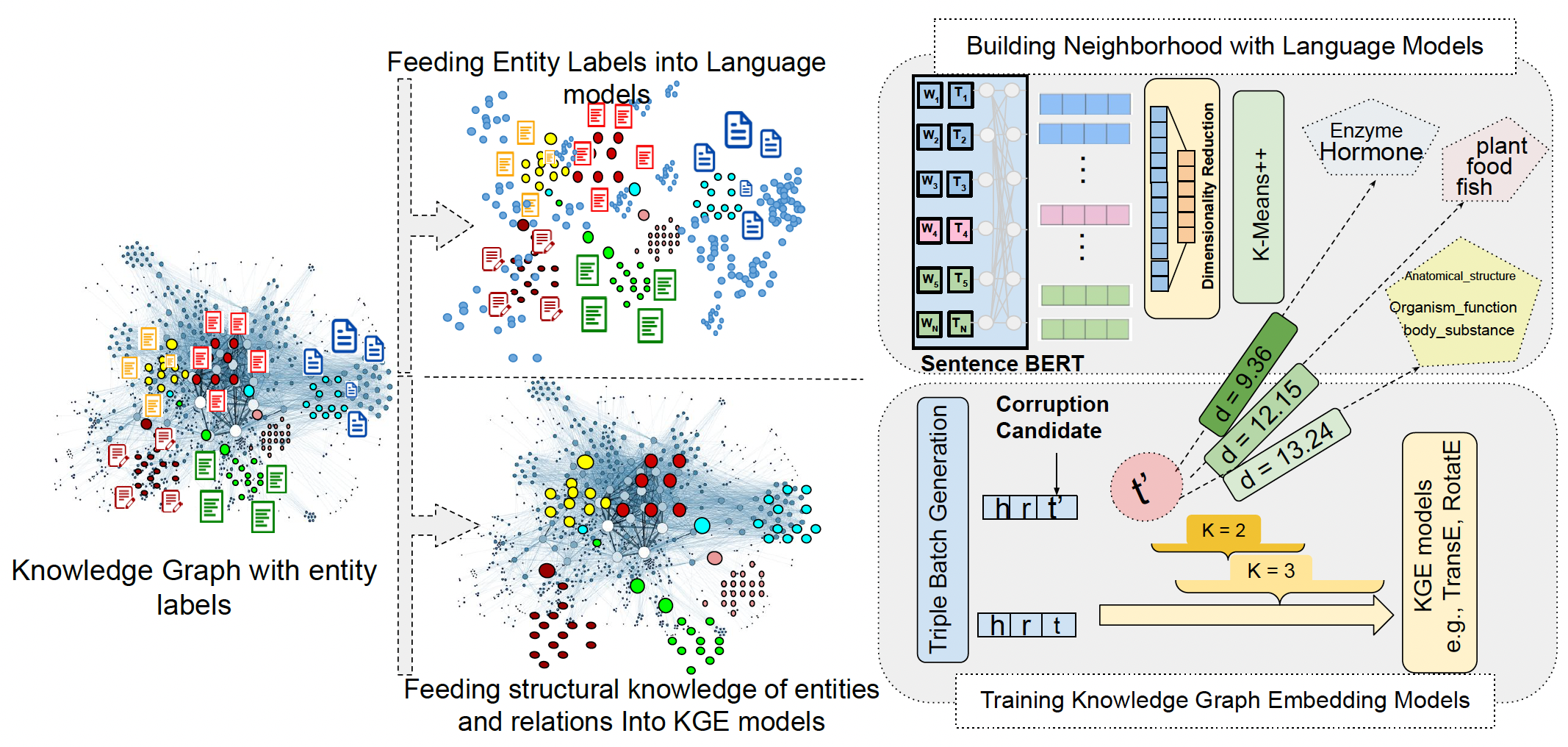}
    \caption{KGE with LEMON. In Sentence BERT part of the figure, $w_{i}\in\mathcal{E}_{text}$ and $T_{j}\in w_{i}$, where $w_{i}$ indicates a word in the entity text $\mathcal{E}_{text}$ and $T_{j}$ represents the tokenized chunk of the word $w_{i}$.}
    \label{fig: neigborhood_clustering}
\end{figure*}

%This sub-phase of our execution demonstrates the construction of the neighborhood clusters. 
Building the neighborhood is considered a crucial step, since it puts the KG entities in different clusters based on the pre-trained embeddings of the text. Figure~\ref{fig: neigborhood_clustering} illustrates our proposed neighborhood clustering step. 

%In order to build the clusters consisting of member entities, output embeddings from sentence transformer is required. 
The input for Sentence-BERT is the textual representation of the set  $\mathcal{E}_{text}$ which is the set of corresponding textual representation of the entity set $\mathcal{E}$.  
% Lets consider it as $\mathcal{E}_{text}$. 
Let us consider Sentence-BERT as a function $\Omega$. The output of $\Omega$ is a $|\mathcal{E}|\times786$
dimensional %vector 
matrix representation. % where $|\mathcal{E}|$ is the number of entities in $\mathcal{E}$, which is stated in equation \ref{eq: st_output}. 
Formally, this can be defined as:

\begin{equation}
\label{eq: st_output}
    \Omega:\mathcal{E}_{text}\to\mathcal{E}_{ST}\in\mathcal{R}^{|\mathcal{E}|\times786}
\end{equation}

The "curse of dimensionality" refers to the fact that large dimensional embeddings can be harmful to learning systems~\cite{verleysen2005curse}. Because PLM embeddings are large, we suspect that, it may not be a suitable fit for clustering them for the link prediction challenge. To mitigate the issue, we have used dimensionality reduction algorithm to reduce the PLM embedding dimension to $\mathcal{Z}$ with a value of less than 768 in order to avoid the curse of dimensionality. 
Let us define the dimensionality reduction function as $\mathcal{\phi}$. The input to this function is the output of $\Omega$ which is defined in equation~\ref{eq: st_output}. %Additionally $\phi$ requires the desired number of reduced dimension $\mathcal{Z}$ as a hyperparameter. 
The output of $\phi$ is a $|\mathcal{E}|\times\mathcal{Z}$ dimensional vector namely $\mathcal{E}_{ST}^{\mathcal{Z}}\in\mathcal{R}^{|\mathcal{E}|\times\mathcal{Z}}$. Formally, this can be defined as:
 {\LinesNumberedHidden
 \begin{algorithm}
\caption{Building neighborhood.}
    %\tcp{Below Algorithm is for generating grounding in the pipeline}
    \label{algorithm: Algorithm_1}
    \begin{flushleft}
        \textbf{INPUT:} Entity embedding $\mathcal{E}$ from PLMs, Desired number of clusters $\mathcal{K}$, dimensionality reduction function $\mathcal{\phi}$, reduced dimensionality $\mathcal{Z}$. 
         ~\\  \textbf{OUTPUT:} Cluster Dictionary dict where~\\keys$\gets$ mapped entities\{e\}$\subseteq\mathcal{E}$~\\values$\gets$ cluster centroids \{c\},~\\ $KM_{\theta}$ as K-means++ Attributes such as distance to centroids, number of centroids etc.~\\
    \end{flushleft}
    \SetKwFunction{FMain}{Build entity clusters}{}
    \SetKwProg{Pn}{Function}{:}{}
    \Pn{\FMain{$\mathcal{E},\mathcal{K}, \mathcal{Z}$}}{
         $\mathcal{E}_{\mathcal{Z}} \gets \mathcal{\phi}(\mathcal{E}, \mathcal{Z})$~\\
         KM $\gets$ Initialize K-means $(number of clusters = \mathcal{K})$~\\
        $D_{\mathcal{E}_{\mathcal{Z}}}, \mathcal{C}, KMeans\_attributes \gets KM(\mathcal{E}_{\mathcal{Z}})$~\\
        \tcp{For each entities in $\mathcal{E}_{\mathcal{Z}}$ obtain assigned cluster number $\mathcal{C}$}
        \tcp{$D_{\mathcal{E}_\mathcal{Z}}$ is the distance between cluster centroids}
        $dict$ $\gets$ \{\}~\\
        \For{each entity $e$ $\in$ $\mathcal{E}$}{
            $dict\{key:e, value: c\}$ $\gets$ map(\{$e$\}$\subseteq\mathcal{E}_{\mathcal{Z}}$, $c\in\mathcal{C}$)~\\  
            \tcp{map the matching entities \{$e$\}$\subseteq\mathcal{E}$ belonging to the centroid $c\in\mathcal{C}$}
        }
       \textbf{Return} $dict$, $KM_\theta$
        }
 \end{algorithm}}

\begin{equation}
\label{eq: dr_output}
    \phi:\mathcal{E}_{ST}\to \mathcal{E}_{ST}^{\mathcal{Z}}\in\mathcal{R}^{|\mathcal{E}|\times\mathcal{Z}}
\end{equation}

K-means++ is applied on the resultant entity vectors from PLMs $\mathcal{E}_{ST}^{\mathcal{Z}}$ to obtain $\mathcal{K}$ clusters consisting of entities $\{e\}\subseteq\mathcal{E}$. Other K-means attributes such as cluster centroids and distances between the cluster centroids are retrieved as well. The objective function for building the neighborhood is aligned with K-means clustering algorithms' objective, formally defined as:

\begin{equation}
\label{eq: loss_function_kmeans}
    \mathrm{arg min}_{\mathcal{C}}^{}\sum_{i=1}^{\mathcal{K}}\sum_{e_i\subseteq\mathcal{E}_{ST}^{\mathcal{Z}}\in\mathcal{C}}^{|\mathcal{E}|}\left\| \mathcal{E}_{ST}^{\mathcal{Z}}-\mu_i \right\|^2
\end{equation}

Here, $\mu$ is a variable, optimized for each cluster centers $c_{i}$.  Finally, each entities that are mapped to their respective cluster centroids acts as the representative of that particular cluster. After assigning each entity $e$ to its respective clusters $c_i$ $\in$ $\mathcal{C}$,  we construct the mapping dictionary $dict$, where the cluster assignment of each entity is preserved. In $dict$, the keys are the set of all the entity symbols $\mathcal{E}$ and the values are associated to their representative cluster centroids $\mathcal{C}$ (Equation~\ref{eq:dictionary}). The distances of the cluster centroids are preserved under the attribute of K-means++, namely $KM_\theta$.% That is how basically our neighborhood cluster is built to be used in training. 
 
 \begin{equation}
\label{eq:dictionary}
    dict:\mathcal{E}\to\mathcal{C}
\end{equation}

 The neighborhood building process is summarized in Algorithm~\ref{algorithm: Algorithm_1}.
 
 % ALGO HERE 

\subsection{Negative Sample Generation}

Typically, KGE models create negative samples from the existing positive triples during the training process. Either the head or the tail is corrupted by replacing it with candidate entities which eventually form negative triples. We follow the standard way to generate negative samples during the training phase. %Previous subsection (subsection \ref{sub: building the neighborhood}) talks about the neighborhood creation, which is used in this phase as the mean of generation of the negative samples. 
However, our proposed method utilizes the created cluster dictionary $dict$ and K-means attributes $KM_\theta$ to generate negative samples. Our proposed approach aims to generate negative samples while preserving the contextual meaning of the entity, so that meaningful negative samples are fetched. Our intuition is that, KGE models may exhibit better performance, if meaningful entities are selected while the corruption process.

%The cluster dictionary $dict$ and K-means attributes $KM_\theta$ from the subsection \ref{sub: building the neighborhood} is required regarding the information of the entities and their associated clusters.
Let us define the training set as $\mathcal{T}_{h,r,t}$, where \textit{h},\textit{r},\textit{t} represents head, relation, and tail, respectively. The %stored information about the 
K-means++ attributes, $KM_{\theta}$  %(information about the cluster centroids and distances between them)
are used to obtain the distance between candidate entities to be corrupted and the cluster centroids. Our goal is to generate $\mathcal{N}$ negative samples per positive triple $t_{h,r,t}$ in each given batch $\mathcal{B}$. We only consider the entities $\{e\}\subseteq\mathcal{E}$, that are within $\mathcal{H}$ hops and distance $d$. Assuming that within $\mathcal{H}$ nearest hops we have $k\in\mathcal{K}$ clusters. For each triple $t_{h,r,t}$, firstly the head or tail corruption is done in a probabilistic manner. For the target entity $e$, we compute the distances from the centroids of each cluster. %From the neighborhood building function (discussed in Subsection \ref{sub: building the neighborhood}), we have 
The information of the distance between clusters is obtained from the $dict$ which has been obtained in the phase of building the neighborhood (algorithm ~\ref{algorithm: Algorithm_1}). The clusters are then sorted based on their distances $d$ which is between the entity to be corrupted and the other cluster centroids. Let us define the distance function as $CD$, the cluster centroids that are stored in K-means attribute $KM_\theta$ as $C$, the desired number of clusters as $\mathcal{K}$, and obtained distance vector is $M$, formally defined as: %then the formulation (equation \ref{eq: distance_function}) can be considered as follows.

\begin{equation}
\label{eq: distance_function}
    CD(e,\mathcal{C}_{KM_\theta})= M\in\mathcal{R}^{1\times|\mathcal{K}|}
\end{equation}

Upon obtaining the sorted clusters and their member entities, it is possible to obtain $\mathcal{N}$ negative entities randomly as the counterpart for the target entity $e$. With the randomly picked entities $\{e'\}$ from the hop $\mathcal{H}$ clusters within the distance $d$, the negative triple $\mathcal{T}_{h',r,t'}$ is formed. Let us consider obtaining $\mathcal{N}$ negative samples $\{e_0', e_1',...e_{\mathcal{N}}'\}$ for a particular entity and the number of clusters based on tolerated distances to clusters as $d_{max}$. The set of possible negative entities resides in the union of all the clusters till that distance $d_{max}$. We can formally define that as:

\begin{equation}
\label{eq: negative_sample_set}
  \{e_0', e_1',...,e_{\mathcal{N}}'\}\in\mathcal{E'}=\mathcal{E}_{H_0}\cup\mathcal{E}_{H_1}\cup\ ...\mathcal{E}_{H_{d_{max}}}  
\end{equation}

The whole process of obtaining the negative sampling during the training process can be described briefly in algorithm \ref{algorithm: Algorithm_2}. 
In algorithm \ref{algorithm: Algorithm_2}, for each triple $t_{h,r,t}$ in $\mathcal{T}_{batch}$ we first decide about the head or tail corruption and apply the corruption on either the head or tail position. The entity to be corrupted is then compared with other cluster centroids. The information about the clusters centroids and their member entities is available in the built neighborhood which is represented as a dictionary $dict$. It is later sorted based on the distance between the entity to be corrupted namely $entity2corrupt$ and its distance to other cluster centroids. We consider fetching the entities for the corruption from only the first $\mathcal{H}$ clusters from $dict_{sorted}$. The $\mathcal{N}$ corrupted entities are then picked from these nearest clusters in a random fashion. The negative triple is formed by replacing the candidate or target entity in the corrupt position with the chosen $corruption\_candidates$ which gives us $\mathcal{N}$ corrupted triples $t'$ aligned in the current batch.      
  % ALGO HERE 

\begin{algorithm}[h]
\caption{LM-Driven Negative Sampling.}
    %\tcp{Below Algorithm is for generating grounding in the pipeline}
    \label{algorithm: Algorithm_2}
    \begin{flushleft}
        \textbf{Input:} Training set $\mathcal{T}_{\textsubscript{h,r,t}}$, cluster dictionary $dict$, K-means attributes $KM_{\theta}$,
        Entity set $\mathcal{E}$, Relation set $\mathcal{R}$, Batch size $\mathcal{B}$, Negative sample number $\mathcal{N}$, Number of nearest $\mathcal{H}$ hops based on $d_{max}$ distance~\\  
        \textbf{Output:}For given batch of triples $\mathcal{T}_{batch}$ generate batch of negatives $\mathcal{T}'_{batch}$ where $h',r,t'\in\mathcal{T'}_{batch}$
    \end{flushleft}
    \SetKwFunction{FMain}{Sample Negative}
    \SetKwProg{Fn}{Function}{:}{}
    %}
    \Fn{\FMain{$\mathcal{T}_{\textsubscript{h,r,t}},d,KM_{\theta},\mathcal{R},\mathcal{B},\mathcal{N},\mathcal{H}$}}{
  $\mathcal{T}'_{batch}$ $\gets$[~]~\\
    \For{triple $t_{h,r,t} \in \mathcal{T}_{batch}$} {
        $corrupt\_position \gets probability(h, t, 0.5)$~\\ %$\Comment{bern negative sampling to decide whether head or tail corruption}~\\
        $entity2corrupt \gets t_{h,r,t}[corrupt\_position]$~\\
        $distance\_to\_clusters$ $\gets$ 
        $compute\_distance$ ($entity2corrupt$, $dict$,
        $KM_{\theta})$~\\
        $dict_{sorted}$ $\gets$ $sort(dict$, $distance\_to\_clusters)$~\\
        $corruption\_candidates \gets dict_{sorted}[0:\mathcal{H}]$~\\
        $corrupted\_entities$ $\gets$ $random\_choice(corruption\_candidates,$~\\ $\mathcal{N})$~\\
        Negative Triples $t': t[corrupt\_position] \gets corrupted\_entities$~\\
        $\mathcal{T}'_{batch} \gets \mathcal{T}'_{batch} \cup t'$
    }
       Return $\mathcal{T}'_{batch}$
     }
\end{algorithm} 

\begin{table*}[t]
\centering
\begin{adjustbox}{width=\textwidth}
\begin{tabular}{l|c|c|c|c|c|c|c|c}
\toprule
\textbf{Dataset} & $|E|$ & $|R|$ 
& \textbf{\#train} & \textbf{\#valid} & \textbf{\#test} & \textbf{Vocabulary} & \textbf{Avg. \#Chars $(E)$} & \textbf{Avg. \#Chars $(R)$}\\\midrule
%\textbf{Nations} & 14 & 55 & 1,992 & 1,592 & 199 & 201 \\
%\textbf{UMLS} & 135 & 46 & 6,529 & 5,216 & 652 & 661 \\
%\textbf{CoDEx-S} & 2,034 & 42 & 36,543 & 32,888 & 1,827 & 1,828\\
%\textbf{CoDEx-M} & 17,050 & 51 & 206,205  & 185,584 &  10,310 & 10,311\\
%\textbf{CoDEx-L} & 77,951 & 69 & 612,437 & 551,193 & 30,622 & 30,622\\\bottomrule
\textbf{WN18} & 40,943  & 18 & 141,442 & 5,000 & 5,000 & 126,087 & 19.691 & 19.000 \\
\textbf{WN18RR} & 40,943 & 11 & 86,835 & 3,034 & 3,134 & 81,954 & 19.691 & 18.455\\
\bottomrule
%\textbf{CoDEx-L} & 77,951 & 69 & 551,193 & 30,622 & 30,622 & 91,918 & 39.294 & 96.754\\\bottomrule
\end{tabular}
\end{adjustbox}
\caption{Dataset statistics. This table presents the number of entities, relation, triples as well as division of train, validation and test sets, where \textit{Chars} refers to characters on WN18 and WN18RR.}
\label{tab:data}
\end{table*}
%\vspace{-0.5cm}

\section{Experimental Set-up}
% In this section, we provide the all the details of the experimental set up.
% This includes the description of the datasets and the Hyper-parameter Settings, as well as the baseline models used in this work. 
%including the datasets, hyper-parameters, and the chosen KGE models.

\subsection{Dataset}
We evaluate our approach on publicly available datasets of namely:  WN18~\cite{bordes2013translating}, and WN18RR~\cite{dettmers2018convolutional}.
WN18 dataset contains lexical relations between words \cite{sun2018rotate} where several patterns can be found from that dataset including symmetric, anti-symmetric and inverse.
WN18RR is a subset of WN18 in which many of the inverse relations are removed due to their leakage. 
The main patterns in this dataset are: asymmetric, symmetric, and composition~\cite{sun2018rotate}. 
In Table~\ref{tab:data}, we provide the statistical information for number of entities, relations, triples, vocabulary and average characters of these datasets. 
FB15k-237 is not used due to the missing text or missing entity labels. As a result, many of the triples are to be removed from the dataset and KGE models are unable to perform link prediction in a desired way.

\subsection{Hyper-parameter Settings}
For WN18 and WN18RR many of the hyperparameters are taken from the best settings of RotatE~\cite{sun2018rotate}, also collected from their Github~\footnote{\url{ https://github.com/DeepGraphLearning/KnowledgeGraphEmbedding}}. 
For these two datasets, both adversarial and non-adversarial methods~\cite{sun2018rotate} have been considered. 
For the underlying KGEs, the embedding dimension $\mathcal{D}$=\{400, 500, 1000\}, the margin $\gamma$=\{6.0, 12.0\}, learning rate $\alpha$=\{0.0001, 0.00005, 0.001, 0.002\}, the temperature $\tau$=\{0.5, 1.0\}, the number of NS $\mathcal{N}$=\{50,100\}, batch size $\mathcal{B}$=\{512, 1024\} were used. 
The specific hyperparameters of our approach are $\mathcal{K}$, $\mathcal{H}$ in the range of \{10,20\} and \{2,3,5\}, accordingly which correspond to the number of clusters and number of $d$ distant hops. 
The dimensionality of the embedding from PLM has been reduced to 200 after performing a number of experiments and inital analysis. 

\subsection{Baseline Models and Evaluation Metrics}
We evaluated our proposed approach on the following baseline models: 1) \textbf{RotatE} - a popular KGE model trained with the score function $\left\Vert h \circ r-t \right\Vert$, \textbf{TransE}~\cite{bordes2013translating} with the score function $\left\Vert h+r-t \right\Vert$, and \textbf{DistMult} is a KGE model that employs a bi-linear formulation with the score function of  $\left\Vert h * r*t \right\Vert$.

\begin{table*}[htb!]
\begin{adjustbox}{width=1\textwidth}
\begin{tabular}{c|l|c|cccccc} 
\toprule
\multicolumn{1}{l}{}                & \multicolumn{1}{l}{} &        & \multicolumn{6}{c}{\textbf{Without Adverserial}}                                                                                                                                                                           \\ 
\hline
\multicolumn{1}{l|}{\textbf{Model}} & Datasets             & Metric & KBGAN                                  & NSCaching                              & Uni.                                  & Uni. SANS                              & Uni. RW-SANS                          & LEMON           \\ 
\hline
\textbf{Rotate}                     & WN18                 & MRR    & -                                      & -                                      & 0.9474                                & 0.9499                                 & 0.9489                                & 0.9481          \\
                                    &                      & H@10   & -                                      & -                                      & 96.09                                 & 95.97                                  & 96.07                                 & \textbf{96.18}  \\
                                    &                      &        & -                                      & -                                      & {\cellcolor[rgb]{0.702,1,0.702}}+0.09 & {\cellcolor[rgb]{0.702,1,0.702}}+0.21  & {\cellcolor[rgb]{0.702,1,0.702}}+0.11 & -               \\ 
\cline{2-9}
                                    & WN18RR               & MRR    & -                                      & -                                      & 0.4711                                & 0.4769                                 & 0.4796                                & 0.4732          \\
                                    &                      & H@10   & -                                      & -                                      & 56.51                                 & 55.76                                  & 57.12                                 & \textbf{57.33}  \\
                                    &                      &        & -                                      & -                                      & {\cellcolor[rgb]{0.702,1,0.702}}+0.82 & {\cellcolor[rgb]{0.702,1,0.702}}+1.57  & {\cellcolor[rgb]{0.702,1,0.702}}+0.21 & -               \\ 
\hline
\textbf{DistMult}                   & WN18                 & MRR    & 0.7275                                 & 0.8306                                 & 0.4689                                & 0.7553                                 & 0.6235                                & 0.4786          \\
\multicolumn{1}{l|}{}               &                      & H@10   & 93.08                                  & 93.74                                  & 81.39                                 & 93.19                                  & 89.80                                 & 82.93           \\
                                    &                      &        & {\cellcolor[rgb]{1,0.851,0.851}}-10.15 & {\cellcolor[rgb]{1,0.851,0.851}}-10.77 & {\cellcolor[rgb]{0.702,1,0.702}}+1.54 & {\cellcolor[rgb]{1,0.851,0.851}}-10.26 & {\cellcolor[rgb]{1,0.851,0.851}}-6.87 & -               \\ 
\cline{2-9}
\multicolumn{1}{l|}{}               & WN18RR               & MRR    & 0.2039                                 & 0.4128                                 & 0.3938                                & 0.4025                                 & 0.4071                                & 0.3953          \\
\multicolumn{1}{l|}{}               &                      & H@10   & 29.52                                  & 45.45                                  & 52.86                                 & 44.74                                  & 49.09                                 & \textbf{52.88}  \\
                                    &                      &        & {\cellcolor[rgb]{0.702,1,0.702}}+23.36 & {\cellcolor[rgb]{0.702,1,0.702}}+7.43  & {\cellcolor[rgb]{0.702,1,0.702}}+0.02 & {\cellcolor[rgb]{0.702,1,0.702}}+8.14  & {\cellcolor[rgb]{0.702,1,0.702}}+3.79 & -               \\ 
\hline
\textbf{TransE}                     & WN18                 & MRR    & 0.6606                                 & 0.7818                                 & 0.6085                                & 0.8228                                 & 0.8195                                & 0.7646          \\
\multicolumn{1}{l|}{}               &                      & H@10   & 94.80                                  & 94.63                                  & 95.53                                 & 95.09                                  & 95.22                                 & \textbf{95.57}  \\
                                    &                      &        & {\cellcolor[rgb]{0.702,1,0.702}}+0.77  & {\cellcolor[rgb]{0.702,1,0.702}}+0.94  & {\cellcolor[rgb]{0.702,1,0.702}}+0.04 & {\cellcolor[rgb]{0.702,1,0.702}}+0.48  & {\cellcolor[rgb]{0.702,1,0.702}}+0.35 & -               \\ 
\cline{2-9}
\multicolumn{1}{l|}{}               & WN18RR               & MRR    & 0.1808                                 & 0.2002                                 & 0.2002                                & 0.2254                                 & 0.2317                                & 0.2145          \\
\multicolumn{1}{l|}{}               &                      & H@10   & 43.24                                  & 47.83                                  & 49.63                                 & 51.15                                  & 53.41                                 & 51.22           \\
                                    &                      &        & {\cellcolor[rgb]{0.702,1,0.702}}+7.98  & {\cellcolor[rgb]{0.702,1,0.702}}+3.39  & {\cellcolor[rgb]{0.702,1,0.702}}+1.57 & {\cellcolor[rgb]{0.702,1,0.702}}+0.07  & {\cellcolor[rgb]{1,0.851,0.851}}-2.19 & -               \\
\bottomrule
\end{tabular}
\end{adjustbox}
\caption{Performance of algorithms without adversarial NS. Here, \textit{Uni.} is the short form of the algorithm \textit{Uniform} and $\Delta$ indicates the difference between LEMON and the other NS approaches on Hit@10. The green color shows the increase and the red color is dedicated for decreases in performance comparisons."-" indicates the unavailability.}
\label{tab:result_without_adv_WN}
\end{table*}

\begin{table}[htb!]
\centering
\begin{adjustbox}{width=0.50\textwidth}

\begin{tabular}{c|l|c|cccc} 
\toprule
\multicolumn{1}{l}{}                & \multicolumn{1}{l}{} &        & \multicolumn{4}{c}{\textbf{With Adverserial}}                                                                                              \\ 
\hline
\multicolumn{1}{l|}{\textbf{Model}} & Datasets             & Metric & Uni.                                  & SANS                                   & RW-SANS                                & LEMON            \\ 
\hline
\textbf{Rotate}                     & WN18                 & MRR    & 0.9498                                & 0.9494                                 & 0.9496                                 & 0.9469           \\
                                    &                      & H@10   & 96.05                                 & 95.85                                  & 96.09                                  & 95.93            \\
                                    &                      &        & {\cellcolor[rgb]{1,0.851,0.851}}-0.12 & {\cellcolor[rgb]{0.702,1,0.702}}+0.08  & {\cellcolor[rgb]{1,0.851,0.851}}-0.16  & -                \\ 
\cline{2-7}
                                    & WN18RR               & MRR    & 0.4760                                & 0.4745                                 & 0.4805                                 & \textbf{0.4770}  \\
                                    &                      & H@10   & 57.29                                 & 57.12                                  & 56.94                                  & \textbf{57.35}   \\
                                    &                      &        & {\cellcolor[rgb]{0.702,1,0.702}}+0.06 & {\cellcolor[rgb]{0.702,1,0.702}}+0.23  & {\cellcolor[rgb]{0.702,1,0.702}}+0.41  & -                \\ 
\hline
\textbf{DistMult}                   & WN18                 & MRR    & 0.6837                                & 0.7561                                 & 0.6634                                 & 0.6824           \\
\multicolumn{1}{l|}{}               &                      & H@10   & 92.94                                 & 93.04                                  & 91.08                                  & 92.90            \\
                                    &                      &        & {\cellcolor[rgb]{1,0.851,0.851}}-0.04 & {\cellcolor[rgb]{1,0.851,0.851}}-0.14  & {\cellcolor[rgb]{0.702,1,0.702}}+1.82  & -                \\ 
\cline{2-7}
\multicolumn{1}{l|}{}               & WN18RR               & MRR    & 0.4399                                & 0.3684                                 & 0.3836                                 & 0.4374           \\
\multicolumn{1}{l|}{}               &                      & H@10   & 53.80                                 & 38.70                                  & 42.74                                  & 52.80            \\
                                    &                      &        & {\cellcolor[rgb]{1,0.851,0.851}}-1.00 & {\cellcolor[rgb]{0.702,1,0.702}}+14.10 & {\cellcolor[rgb]{0.702,1,0.702}}+10.06 & -                \\ 
\hline
\textbf{TransE}                     & WN18                 & MRR    & 0.7722                                & 0.7136                                 & 0.7429                                 & \textbf{0.7947}  \\
\multicolumn{1}{l|}{}               &                      & H@10   & 92.02                                 & 84.06                                  & 88.51                                  & \textbf{94.92}   \\
                                    &                      &        & {\cellcolor[rgb]{0.702,1,0.702}}+2.90 & {\cellcolor[rgb]{0.702,1,0.702}}+10.86 & {\cellcolor[rgb]{0.702,1,0.702}}+6.41  & -                \\ 
\cline{2-7}
\multicolumn{1}{l|}{}               & WN18RR               & MRR    & 0.2232                                & 0.2249                                 & 0.2273                                 & 0.2262           \\
\multicolumn{1}{l|}{}               &                      & H@10   & 52.78                                 & 53.21                                  & 53.81                                  & 53.54            \\
                                    &                      &        & {\cellcolor[rgb]{0.702,1,0.702}}+0.76 & {\cellcolor[rgb]{0.702,1,0.702}}+0.33  & {\cellcolor[rgb]{1,0.851,0.851}}-0.27  & -                \\
\bottomrule
\end{tabular}
\end{adjustbox}
\caption{Performance of algorithms with adversarial NS. \textit{Uni.} denotes \textit{Uniform} and $\Delta$ indicates the difference between LEMON and the other NS approaches on Hit@10. The green and red colors are for increase/decreases in performance comparisons, "-" for not available.}
\label{tab:result_with_adv_WN}
\end{table}

\textbf{Evaluation Metrics}
We follow the baseline models and assess the performance of the KGE models based on different negative sampling methods, with standard metrics such as Hit@10 and Mean Reciprocal Rank (MRR) as reported in \cite{sun2018rotate}, \cite{bordes2013translating} and \cite{ali2021bringing}.
For the evaluation of the KGE models, we rely on link prediction task where the queries are formed as head prediction ($?,r,t$) and tail prediction ($h,r,?$) for an existing true triple in the test set. The $?$ is replaced by all the entities in the Knowledge Graph in order to generate the false triples. The rank of the true triple is considered as the position of it when sorted based on the plausibility against all the possible combinations of corrupted ones. The mean rank (MR) is the average rank of all the test triples in the test set. The lower score in MR indicates better performance.
MRR represents the inverse of the average rank obtained from the score of each test triple. In MRR, a better performance is indicated by a higher value. The Hit@10 represents the accuracy that the target entity for a particular query appeared in the top 10 prediction during the inference. The average Hit@10 score is used to measure the performance on the test set.

\begin{figure*}
\centering\includegraphics[scale=0.56]{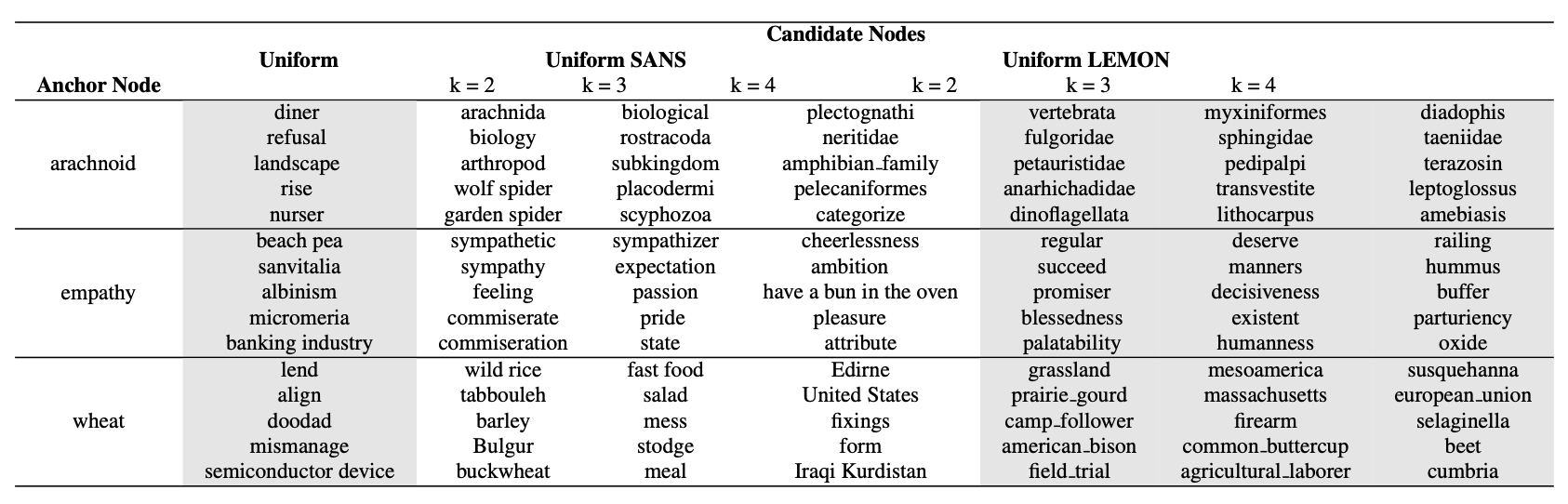}
   \caption{Example set of candidate entities that form negative samples. The Uniform and SANS negative entities are taken from ~\cite{ahrabian-etal-2020-structure}.}
   \label{fig:table}
%  \centering
\end{figure*}

\section{Evaluation and Analysis}
In this section, we present the results of our evaluation on the the two benchmark datasets. 
We compare against the baseline approaches reported in \cite{ahrabian-etal-2020-structure} which includes KBGAN~\cite{cai-wang-2018-kbgan}, NSCaching~\cite{zhang2019nscaching} and SOTA NS approaches for Self Adversarial Negative Sampling~\cite{sun2018rotate}.

\subsection{Evaluation Results} %The results of the evaluations show improvements in the performance metrics specially for Hit@10.
 Table~\ref{tab:result_without_adv_WN} and table~\ref{tab:result_with_adv_WN} reports the results of LEMON and the baseline models, with and without adversarial settings~\cite{sun2018rotate} for NS. %The results are presented for uniform, and adversarial negative sampling.
%In table \ref{tab:result_with_without_adv_WN}, the results are shown without adversarial sampling.
For RotatE KGE, the achieved Hit@10 by LEMON on on WN18 and WN18RR are 96.18\% and is 57.33\%, respectively. 
On WN18, with our proposed negative sampling method, the TransE model achieved a Hit@10 score of 95.57\%. 
Furthermore, a significant improvement in the performance of TransE is visible in both MRR (95.90\%) and Hit@10 (94.92\%) metrics, using LEMON. 
For the RotatE model on WN18RR, we observed an improved result in Hit@10. 
As can be seen, we calculated the $\Delta$ as the difference of performance in Hit@10 of LEMON with other NS methods.
In all the cases, LEMON outperforms other NS approaches used with RotatE, TransE both on WN18, and WN18RR in non adversarial settings (table~\ref{tab:result_without_adv_WN}). 
For the DisMult model, we outperform others on WN18RR in non adversarial settings. 
Some of the large gains in Hit@10 can be observed in table~\ref{tab:result_without_adv_WN} in WN18RR for DistMult and TransE model. 
On WN18RR, the improvement against KBGAN, Uniform Negative Sampling, uni-SANS and RW-SANS are +23.36, +7.43, +8.14 and +3.79, respectively.
On WN18RR, some of the significant achievements are seen on TransE model against KBGAN and NSCaching approaches which are +7.98 and 3.39 respectively.
For achieving comparable results on WN18, a more extensive hyper-parameter search can be deployed which is among our future work.
On the MRR metric, LEMON is getting comparable results. 
We shall note that, the results of the baseline models are reported from the original work on structure aware negative sampling~\cite{ahrabian-etal-2020-structure}. 
In Table~\ref{tab:result_with_adv_WN}, similar improvement in Hit@10 can be seen on WN18 and WN18RR datasets. There is a significant gain in Hit@10 for DistMult model compare to LEMON with SANS(+14.10) and RW-SANS(+10.06). 
TransE model also obtains significant improvement when comparing it with Uniform, SANS and RW-SANS with LEMON on WN18 dataset.
The improvement in Hits@10 are +2.90, +10.86 and +6.41 respectively.

\subsection{Relevant Negative Sampling}
Figure~\ref{fig:table} reports the examples which are drawn as effective candidates for corruption by LEMON.
In this regard, a comparison between LEMON, Uniform and SANS on the WN18RR is shown.
%is the validity test of resultant negative entities that are drawn in the cases of Uniform negative sampling, SANS and LEMON. 
The output of the baseline NS methods are taken from the SANS paper~\cite{ahrabian-etal-2020-structure}. % (results of  Uniform and SANS are taken from), we also used WN18RR dataset for this purpose. 
%The results show the effect of PLM in generating more meaningful NS. 
From the investigation, we observed that the contextual relevance decays as we go into the further hops. 
The analysis of the drawn examples show that meaning-wise, both LEMON and Uniform SANS provide more relevant candidates than Uniform.
This is while Uniform SANS needs a bigger pre-processing time for building neighborhoods. 
Unlike LEMON, it does not consider any complementary knowledge. 
Several ablation studies and comprehensive analysis of the performance and plausibility of the proposed method have been conducted which can be found in the Appendix.
% Similar to SANS, our method is also capable of generating meaningful negative samples. 
% Semantic meaningfulness declines as we include distant hops.

\begin{comment}

\subsection{Influence of Textual Information}
The textual information represented by entities and relation is also varies in terms of information. 
Often the quantity of the text i.e, number of characters, words often does not play a role for the proper representation from the language models, specially for downstream task such as link prediction. The quantitative information of the benchmark datasets is reported in Table~\ref{tab:data}. We observed that, for the structure dominant KGs the quantity of the text matters less. When the structure also corresponds well to the meaning of the text then we conclude the entities are in a better quality to be used as a pre-trained input for the KGE model.

\end{comment}

\section{Conclusion}
%\newpage
% Generation of negative sampling plays a significant role for the performance of KGE models. The impact on the performance is obvious when the obtained negatives are fetched in a systematic way. 
% Many systematic negative samples require a huge computational burden either on the preprocessing or in the negative sampling generation side. Emphasising on the computational burden, our model is more simpler and yet efficient. 
This paper addresses the problem of creating meaningful negative sampling for knowledge graph embedding models.  
We presented LEMON, that replaces the old paradigm of creating negative sampling using a random distribution technique.
%LEMON leverages rich textual complementary knowledge of the entities to train the knowledge graph embedding models. To achieve this, a pre-trained language model is employed along with informative clustering approach.
LEMON is model-independent and easy to integrate into widely used KGE models (i.e., TransE, RotatE and DistMult) and evaluated on standard benchmark knowledge graphs, namely WN18 and WN18RR. 
The empirical results exhibit a significant improvement in the performance of the underlying KGE models. 
Some of the biggest gains of our approach can be seen in DistMult model against KBGAN, Uniform Negative Sampling, Uni-SANS which are +23.36, +7.43 and +8.14 respectively on WN18RR dataset considering non adversarial settings. 
%LEMON largely outperforms KBGAN and NSCatching for TransE model on WN18RR dataset by obtaining a leap of +7.98 and +3.39 respectively in the same settings in Hit@10. 
In the adversarial settings significant improvement over SANS and RW-SANS can be seen on WN18RR dataset for DistMult model which are +14.10 and +10.06 respectively.
%In the same settings for TransE we have obtained a performance gain of +10.86 and +6.41 is obtained on WN18 dataset. 
%Additionally, we performed clustering tasks and the results approve that using language models-driven negative sampling highly affects in improving performance of KGE models. 
In future work, we plan to consider other language models and combine LEMON with other KGE models. 
% \enlargethispage{100pt}

\section*{Limitations}
The embedding vectors obtained from PLMs are highly important for LEMON. 
The generation of negative samples considers the quality of the contextual representation of the entity text in order to fetch the relevant and useful negative candidates for corruption. 
%Often obtaining such useful text requires to pass through a complicated procedure. 
On the other hand, the structure of the underlying Knowledge Graphs often does not align with the the contextual meaning of the text. 
In order to align them properly, a quality check might be needed for the input text of PLMs.

%\clearpage

%This is essential for obtaining the neighborhood based on clusters in our approach.      

%The proposed approach highly depends on the quality of textual information. There are knowledge graphs in which such complementary knowledge is completely missing.
%In order to make use of language models, the textual information needs to be collected through a complex procedure. 
%Additionally, for the knowledge graph with textual information needs to go through a complex data quality check. Our model falls short when the dissimilarity between the structural and textual information is huge. 

\bibliography{anthology,custom}
\bibliographystyle{acl_natbib}

\appendix

\section{Appendix}
\label{sec:appendix}

%\subsection{Example of a tail corruption}
%Figure \ref{fig:exampel} demonstrates the difference between our approach and Random negative sampling approach. In order to corrupt a particular tail entity (in this case $Vitamin$), the distance between the candidate entity and the clusters (containing entities) plays role. The box in the upper portion of figure \ref{fig:exampel} which demonstrates the distances between the the clusters based on various distances $d=\{9.36, 12.15, 13.24\}$ from the candidate entity $Vitamin$. Each value of $d$ represents the distance between K th cluster. We choose contextually meaningful entities from the clusters from desired distances. On the other hand, the box below shows the sampling method of random corruption which might not sample meaningful entities. 

\subsection{Datasets}
Apart from WN18~\cite{bordes2013translating}, and WN18RR~\cite{dettmers2018convolutional}, we also trained TransE and RotatE on the following datasets:
Nations~\cite{kok2007statistical}, UMLS~\cite{mccray2003upper}, and CoDEx~\cite{safavi-koutra-2020-codex}

Nations and UMLS are small scale yet efficient benchmarks. Nations includes a set of relationships between nations and their features~\cite{kok2007statistical}. The dataset consists of binary and unary relations while UMLS dataset (standing for Unified Medical Language System ) is a high-level ontology for organizing a large number of terminologies used in the biomedical domain. It translates into a unified vocabulary that allows for uniform access to medical resources.  CoDEx provides three comprehensive knowledge graph datasets that include positive and hard negative triples, entity types, entity, and relation descriptions. The knowledge graph in CoDEx is constructed from Wikidata~\cite{vrandevcic2012wikidata}. 

\begin{table*}[t]
\centering
\begin{adjustbox}{width=\textwidth}
\begin{tabular}{l|c|c|c|c|c|c|c|c}
\toprule
\textbf{Dataset} & $|E|$ & $|R|$ 
& \textbf{\#train} & \textbf{\#valid} & \textbf{\#test} & \textbf{Vocabulary} & \textbf{Avg. \#Chars $(E)$} & \textbf{Avg. \#Chars $(R)$}\\\midrule
%\textbf{Nations} & 14 & 55 & 1,992 & 1,592 & 199 & 201 \\
%\textbf{UMLS} & 135 & 46 & 6,529 & 5,216 & 652 & 661 \\
%\textbf{CoDEx-S} & 2,034 & 42 & 36,543 & 32,888 & 1,827 & 1,828\\
%\textbf{CoDEx-M} & 17,050 & 51 & 206,205  & 185,584 &  10,310 & 10,311\\
%\textbf{CoDEx-L} & 77,951 & 69 & 612,437 & 551,193 & 30,622 & 30,622\\\bottomrule
%\textbf{WN18} & 40,943  & 18 & 141,442 & 5,000 & 5,000 & 126,087 & 19.691 & 19.000 \\
%\textbf{WN18RR} & 40,943 & 11 & 86,835 & 3,034 & 3,134 & 81,954 & 19.691 & 18.455\\
\textbf{Nations} & 14 & 55 & 1,592 & 199 & 201 & 1,992 & 7.786 & 14.455\\
\textbf{UMLS} & 135 & 46 & 5,216 & 652 & 661  & 2,614 & 20.830 & 13.336\\
%\textbf{CoDEx-S} &  2,034 & 42 & 32,888 & 1,827 & 1,828 & 5,492 & 50.165 & 104.357\\
\textbf{CoDEx-M} & 17,050 & 51 & 185,584  & 10,310 &  10,311  & 23,492 & 45.435 & 102.725\\\bottomrule
%\textbf{CoDEx-L} & 77,951 & 69 & 551,193 & 30,622 & 30,622 & 91,918 & 39.294 & 96.754\\\bottomrule
\end{tabular}
\end{adjustbox}
\caption{Dataset statistics. This table presents the number of entities, relation, triples as well as division of train, validation and test sets, where \textit{Chars} refers to Characters of Nations, UMLS and CoDEX-M.}
\label{tab:data}
\end{table*}
%\vspace{-0.5cm}

\subsection{Hyperparameters}
For UMLS, Nations and CoDEx-M datasets we have used TransE, and RotatE model for evaluation of our negative sampling approach. 
For fairness of evaluations, a fixed set of hyperparameters were used. 
For nations and UMLS the $\mathcal{D}$, $\gamma$, $\alpha$, $\mathcal{B}$ and $\mathcal{Z}$ are set to 100, 12.0, 0.01, 64 and 2 respectively. 
The $\mathcal{N}$, $\mathcal{K}$ and $\mathcal{H}$ are kept in range of \{3,10\}, \{4,5,6,7,8\} and \{3,4,5,6,7\} accordingly. 
In CoDEx-M, we used $\mathcal{D}$, $\mathcal{\gamma}$, $\alpha$, $\mathcal{B}$, $\mathcal{N}$ as 500, 9, 0.00005, 512, 50. $\mathcal{K}$ and $\mathcal{H}$ are in the ranges of \{8,12\} and \{7,9\} accordingly.

For executing the experiments with Uniform RW-SANS on Nations, UMLS and CoDEx-M datasets, we set the $-nrw$ hyperparameter to 1000 which indicates number of random walks. 
For Nations, UMLS and CoDEx-M the results are obtained without using negative adversarial sampling.

\subsection{Pre-processing time}
In Table~\ref{Pre-processing performance}, we compare the preprocessing time (in minutes) for building the neighborhood in Uniform SANS, RW-SANS and LEMON. In terms of the preprocessing time, LEMON outperforms the other two by taking only 2.9052 minutes in total. 
Meanwhile, Uniform SANS and RW-SANS takes 88.46, and 215.705 minutes accordingly. 
Although, RW-SANS achieved a comparable performance, depending on the length of the entities and relation it can be really resource hungry. The preprocessing times are computed based on the performance on the CoDEx-M dataset. 

\begin{table}[!ht]
\centering
%\begin{table}
\begin{adjustbox}{width=0.50\textwidth}
\begin{tabular}{ll}
\toprule
\textbf{Methods} & \textbf{Total pre-processing time} \\\hline
\textbf{Uniform SANS} & 88.46 minutes \\\hline
\textbf{RW-SANS} & 215.705 minutes \\\hline
\multirow{3}{*}{\textbf{LEMON}} & Embedding Generation: 2.018 minutes \\
 & Cluster Formation: 0.08 minutes \\
 & Total Time: 2.9052 minutes\\\bottomrule
\end{tabular}
\end{adjustbox}
\caption{Pre-processing performance of LEMON and baseline NS methods.}
\label{Pre-processing performance}
%\end{table}
% \CAPTION{PRE-PROCESSING PERFORMANCE OF LEMON AND BASELINE NS METHODS.}
% \LABEL{PRE-PROCESSING PERFORMANCE}
% \end{table}
% \vspace{-0.7cm}
%\end{wraptable}
\end{table}

\begin{comment}

\begin{figure}
%\centering
  \includegraphics[scale=0.11]{images/example3.png}
   \caption{Lemon vs. Random Distribution}
   \label{fig:exampel}
%  \centering
\end{figure}

\end{comment}

\section{Experimental Result on Nations, UMLS and CoDEx-M}
Table~\ref{tab:KGE models with adv} reports the results of baseline algorithms with KGE models with RotatE and TransE on Nations, UMLS and CoDEx-M datasets. 
We observed a performance gain of the KGE models in most of the cases on benchmark datasets with our proposed approach LEMON proposed method. 
The difference between the performance of LEMON and Uniform is stated as $\Delta_{1}$ which is calculated for Hits@1, 3 and 10. 
$\Delta_{2}$ is corresponding to the difference between the LEMON and Uniform SANS while $\Delta_{3}$ is demonstrating the difference between the LEMON and RW SANS negative sampling. 
A considerable performance increase in different models specially in RotatE (by a margin of +3.86 on Hit@1) and TransE (by a margin of +6.28 on Hit@1) can be observed. 
%On the NATIONS dataset, combination of LEMON as negative sampling with all the models outperform others NS approaches.
In few cases, the performance was just comparable with other NS approaches which is (to our observations) due to the structural dominance attribute of the underling KGs.
This fact affects the adaptability of the NS approaches on the employed KGE models. 
We discuss these aspects further in details.

\begin{table*}[htb!]
\centering
\begin{adjustbox}{width=1.0\textwidth}

\begin{tabular}{c|c|cccc|ccll|cccc} 
\toprule
\begin{tabular}[c]{@{}c@{}}\\ \textbf{Model}\end{tabular} & \textbf{Algorithm} & \multicolumn{4}{c|}{\textbf{Nations}}                        & \multicolumn{4}{c|}{\textbf{UMLS}}                           & \multicolumn{4}{c}{\textbf{CoDEx-M}}                                                                           \\ 
\hline
KGE                                                           & Neg. Samp.  & MRR              & H@1             & H@3   & H@10            & MRR              & H@1             & H@3   & H@10            & MRR             & H@1            & H@3             & H@10                                                      \\ 
\hline
                                                              & Uni.            & 0.6375           & 47.26           & 74.62 & 99.25           & 0.8599           & 74.05           & 97.80 & 99.69           & 0.3914          & 31.67          & 42.64           & 53.13                                                     \\
                                                                                                                                         
\textbf{RotatE}                                               & Uni. SANS       & 0.6672           & 50.24           & 80.09 & 99.25           & 0.8616           & 74.28           & 98.03 & 99.62           & 0.3912          & 31.73          & 42.50           & 52.99                                                     \\
                                                                                            
                                                              & Uni. RW-SANS    & 0.6694           & 50.74           & 78.35 & 99.50           & 0.8691           & 76.02           & 97.80 & 99.62           & 0.4007          & 40.07          & 43.78           & 54.46                                                     \\

                                                              & LEMON        & 0.6474           & 47.26           & 77.11 & \textbf{99.75 } & \textbf{0.8713 } & \textbf{77.91 } & 95.99 & \textbf{99.84 } & 0.3942          & 31.95          & 42.77           & 53.38                                           \\ \cline{2-14}            & $\Delta_1$         &     -       &   \cellcolor{green!30} +0.00      & \cellcolor{green!30}+2.49 & \cellcolor{green!30}+0.50  & - &\cellcolor{green!30} +3.86  &  \cellcolor{red!15}-1.81 & \cellcolor{green!30}+0.11 &     -      &   \cellcolor{green!30}  +0.28       &  \cellcolor{green!30}  +0.13  &  \cellcolor{green!30}  +2.15                                      \\              & $\Delta_2$         &     -       &    \cellcolor{red!15}  -3.48      & \cellcolor{red!15} -2.99 & \cellcolor{green!30}+0.50  & - &\cellcolor{green!30} +3.63  &  \cellcolor{red!15}-2.04 & \cellcolor{green!30}+0.22 &     -      &   \cellcolor{green!30}  +0.22       &  \cellcolor{green!30}  +0.27  &  \cellcolor{green!30}   +0.39               \\                                                                          & $\Delta_3$         &     -       &    \cellcolor{red!15}  -3.48      & \cellcolor{red!15} -1.24 & \cellcolor{green!30}+0.25  & - &\cellcolor{green!30} +1.89  &  \cellcolor{red!15}-1.81 & \cellcolor{green!30}+0.22 &     -      &   \cellcolor{red!15}  -8.12       &  \cellcolor{red!15}  -1.01  &  \cellcolor{red!15}   -1.08                                       \\ 
\hline
                                                              & Uni.            & 0.5688           & 36.31           & 70.14 & 99.50           & 0.5149           & 34.56           & 62.63 & 85.70           & 0.3315          & 26.41          & 35.85           & 46.32                                                     \\
\textbf{TransE}                                               & Uni. SANS       & 0.4235           & 6.71            & 73.38 & 98.75           & 0.7376           & 51.05           & 96.97 & 99.47           & 0.3753          & 30.28          & 40.67           & 51.28                                                     \\
                                                              & Uni. RW-SANS    & 0.4168           & 5.72            & 73.38 & 98.75           & 0.7627           & 55.90           & 96.74 & 99.47           & 0.3871          & 31.10          & 42.26           & 52.82                                                     \\
                                                              & LEMON        & \textbf{0.4425 } & \textbf{11.69 } & 71.89 & \textbf{99.25 } & \textbf{0.7689}           & \textbf{57.33}           & 96.21 & \textbf{99.62}           & 0.3746          & 30.12          & 40.71 & 51.40                                           \\\cline{2-14}
                                                               & $\Delta_1$    &    -        &  \cellcolor{green!30} +5.23  &\cellcolor{red!15} -1.74  &\cellcolor{green!30} +0.50 & - &  \cellcolor{green!30}+5.14 &\cellcolor{green!30} +0.22  &\cellcolor{green!30} +0.46  &     -      & \cellcolor{green!30} +0.03  &\cellcolor{green!30} +0.07  &\cellcolor{green!30} +0.01 \\                                                               & $\Delta_2$         &    -        & \cellcolor{green!30}+4.98        & \cellcolor{red!15}-1.49  & \cellcolor{green!30}+0.50 & - &\cellcolor{green!30} +6.28  & \cellcolor{red!15}-0.76  &\cellcolor{green!30} +0.15  &     -      & \cellcolor{red!15} -0.16    & \cellcolor{green!30}+0.04 & \cellcolor{green!30}+0.12 \\
                                                                                                                              & $\Delta_3$         &    -        &  \cellcolor{green!30}  +5.97     & \cellcolor{red!15}-1.49  & \cellcolor{green!30}+0.50 & - & \cellcolor{green!30}+1.43  & \cellcolor{red!15}-0.53  & \cellcolor{green!30} +0.15 &     -      &  \cellcolor{red!15}-0.98    & \cellcolor{red!15}-1.55 & \cellcolor{red!15}-1.42     \\
\bottomrule
\end{tabular}
\end{adjustbox}
\caption{Result of different KGE models with different negative sampling techniques on Nations, UMLS, CoDEx-M datasets. Here, \textit{Uni.} is the short form of the algorithm \textit{Uniform} and $\Delta$ indicates the difference between LEMON and the baseline algorithms, in terms of Hit@10. $\Delta_{1}$, $\Delta_{2}$, and $\Delta_{3}$ refer to the difference between LEMON and \textit{Uni.}, \textit{Uni. SANS} and \textit{Uni. RW-SANS}, respectively. The green color shows the increase and the red color is dedicated for decreases in performance comparisons. The outperforming cases are shown as bold for LEMON.}
\label{tab:KGE models with adv}
\end{table*}

\subsection{Ablation Study on Time Complexity} 
Table~\ref{tab:complexity} demonstrates the complexity of LEMON and the baseline algorithms in terms of pre-processing, run-time, and space complexity. 
Here, $t$ is the number of GAN parameters, $\mathcal{N}$ denotes the number of negative samples, $\mathcal{B}$ is the batch size,  $E$ is an edge set, $\mathcal{E}$ is the set of entities, $R$ for relation set, and $\mathcal{D}$ represents the embedding dimension. Furthermore, $\mathcal{H}$ is the hops count, $r$ for random walk count, and $\mathcal{Z}$ is a PCA dimensionality reduction parameter.  
In the pre-processing part, LEMON has the same dimensionality reduction complexity of PCA, and K-means. The time complexity of PCA algorithm is $\mathcal{O}({\mathcal{D}_{PLM}}^2 \lvert \mathcal{E} \rvert + {\mathcal{D}_{PLM}}^3)$, where $\mathcal{D}_{PLM}$ indicates the output feature dimension of PLM.  
The time complexity of the K-means++ algorithm is $\lvert \mathcal{E} \rvert \mathcal{K}I\mathcal{Z}$. 
Since we are using a reduced dimension for K-means clustering, the feature dimension is $\mathcal{Z}$. Here, $I$ is the number of iteration and $|\mathcal{E}|$ is the length of the entity set. 
LEMON's run-time complexity is $\mathcal{O}(\mathcal{H}log(\mathcal{H})+\mathcal{B}\mathcal{N})$. %In our training procedure we have a sorting mechanism to 
During the training, LEMON sorts the desired number of Hops $\mathcal{H}$, based on the computed distances. 
The sorting complexity is only added to our approach, comparing to the Uniform negative sampling. 
The space complexity of our proposed method is $\mathcal{O}(\lvert \mathcal{H} \rvert)$,  which only depends on the number of desired hops. 
%It is because LEMON does not consider the entities outside of desired range. 
Other complexity of other NS approaches in Table~\ref{tab:complexity} are taken from SANS~\cite{ahrabian-etal-2020-structure}.  

\begin{table*}[h!]
\centering
\begin{adjustbox}{width=\textwidth}
\begin{tabular}{l|c|c|c}
\toprule
\textbf{Algorithm}                      & \multicolumn{1}{c}{\textbf{Preprocessing Complexity}} & \multicolumn{1}{l}{\textbf{Runtime Complexity}} & \textbf{Space Complexity}  \\ 
\hline
Uniform~\cite{bordes2013translating}  &                $\mathcal{O}(1)$                              &     $\mathcal{O}(\mathcal{B}\mathcal{N})$                                   &  $\mathcal{O}(1)$                 \\
KBGAN~\cite{cai-wang-2018-kbgan}     &        $\mathcal{O}(t)$                                      &  $\mathcal{O}(\mathcal{B}\mathcal{N} + \mathcal{B}\mathcal{D} + \mathcal{b} t)$                                       &   $\mathcal{O}(t)$                \\
NSCaching~\cite{zhang2019nscaching} &                 $\mathcal{O}(1)$                               &          $\mathcal{O}(\mathcal{B}\mathcal{N} + \mathcal{B}\mathcal{E})$                              &      $\mathcal{O}(c|\mathcal{R}||V)$             \\
Self-Adv.~\cite{sun2018rotate}   &            $\mathcal{O}(|\mathcal{E}|)$                                  &   $\mathcal{O}(\mathcal{B}\mathcal{N}+\mathcal{B}\mathcal{D})$                                     &  $\mathcal{O}|\mathcal{E}|$                 \\
Uniform SANS~\cite{ahrabian-etal-2020-structure}                &      $\mathcal{O}(|V|^3)log \mathcal{H}$                                   &           $\mathcal{O}(\mathcal{B}\mathcal{N})$                             &     $\mathcal{O}(|V|^2)$              \\
Self-Adv. SANS~\cite{ahrabian-etal-2020-structure}               &      $\mathcal{O}(|V|^3)log \mathcal{H}$        &       $\mathcal{O}(\mathcal{B}\mathcal{N}+\mathcal{B}\mathcal{D})$                                                &       $\mathcal{O}(|V|^2)$             \\
Uniform RW-SANS~~\cite{ahrabian-etal-2020-structure}               &                   $\mathcal{O}(r\mathcal{H}|V|)$                            &             $\mathcal{O}(\mathcal{B}\mathcal{N})$                           &       $\mathcal{O}(c|V|)$          \\
Self-Adv. RW-SANS~~\cite{ahrabian-etal-2020-structure}             &                    $\mathcal{O}(r\mathcal{H}|V|)$                           &            $\mathcal{O}(\mathcal{B}\mathcal{N}+\mathcal{B}\mathcal{D})$                            &       $\mathcal{O}(c|V|)$          \\ 
\hline
 LEMON & $\mathcal{O}({\mathcal{D}_{PLM}}^2( \lvert \mathcal{E} \rvert + {\mathcal{D}_{PLM}}) + \lvert \mathcal{E} \rvert \mathcal{K}I\mathcal{Z})$ %$\mathcal{O}({\mathcal{D}_{PLM}}^2 \lvert \mathcal{E} \rvert + {\mathcal{D}_{PLM}}^3 + \lvert \mathcal{E} \rvert \mathcal{K}I\mathcal{Z})$ 
& $\mathcal{O}(\mathcal{H}log(\mathcal{H})+\mathcal{B}\mathcal{N})$             &   $\mathcal{O}(\lvert \mathcal{H} \rvert)$ \\\bottomrule               
\end{tabular}
\end{adjustbox}
\caption{Complexity comparison of negative sampling techniques in terms of pre-processing, run-time and space complexity.}
\label{tab:complexity}
\end{table*}

\subsection{Ablation Study on Trained Entity Embedding}
\begin{figure*}[ht!]
\centering
\begin{subfigure}{}
    \includegraphics[trim={2.2cm 1cm 2.2cm 1cm},width=\textwidth]{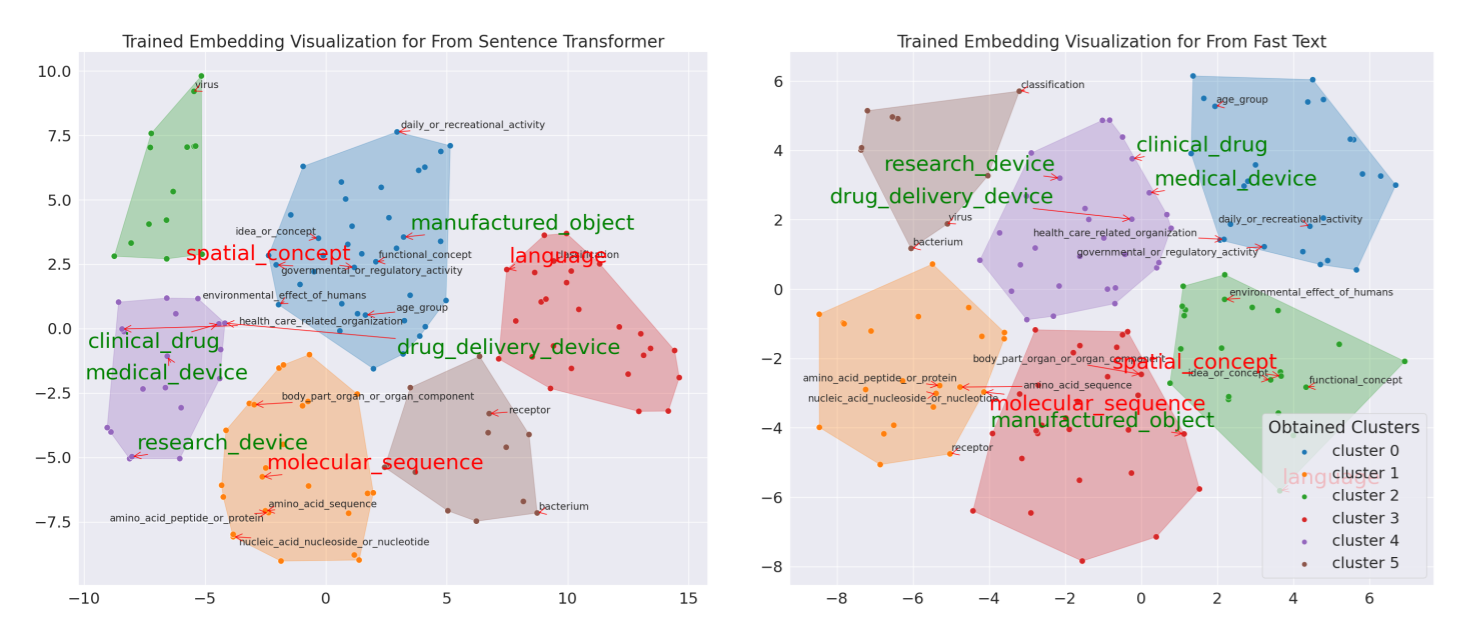}
  \caption{Sentence BERT (left) vs. FastText (right).}
  \label{fig: trained_embedding_PLM}
\end{subfigure}
\begin{subfigure}{}
  \includegraphics[trim={2.2cm 1cm 3cm 0.5cm},width=\textwidth]{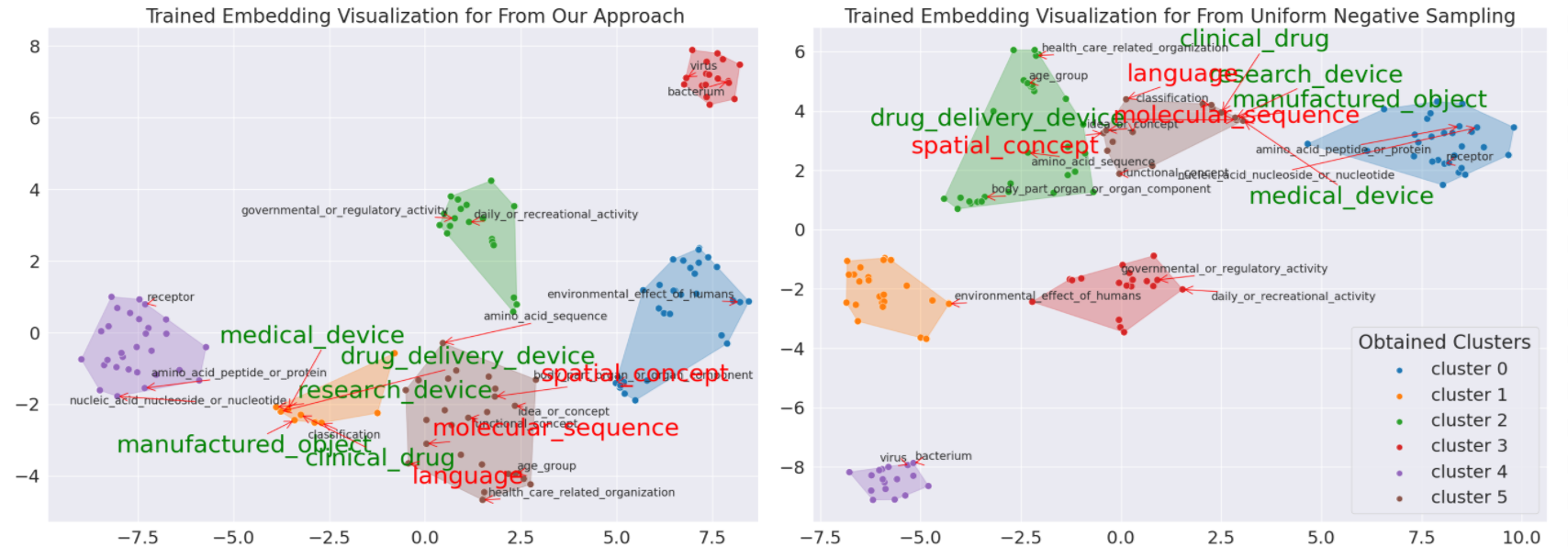}
  \caption{LEMON (left) vs. Uniform NS (right)}
  \label{fig: trained_model_clustering}
\end{subfigure}
\caption[]{(a) Clustering visualization from the embedding vectors obtained from different Sentence Transformer/BERT. (b) Trained model clustering comparing LEMON vs. Uniform NS.}
\end{figure*}

To demonstrate the effectiveness of the language model based clustering in the negative sampling, we illustrate the clusters obtained from the pre-trained embeddings from the Sentence Transformer in Figure~\ref{fig: trained_embedding_PLM}, Sentence BERT in the left and fastText~\cite{bojanowski2017enriching} in the right. 
The clusters depicted in Figure~\ref{fig: trained_embedding_PLM} are constructed based on the UMLS dataset. % the figure is demonstrates the clusters obtained for the entities from UML dataset. 
Through a systematic analysis, the entities which are suppose to be in the same cluster are marked in the same color either green or red.
For example, clinical\_drug, medical\_device, research\_device are green that belong to the same cluster. 
That means the embeddings generated by PLMs are already possessing pre-trained knowledge which can be leveraged by KGEs during the training process. 
Using PLMs has a trade-off as the embeddings generated by them are rich in textual meaning, they lack structural information which can be alleviated by KGE models. 
Figure~\ref{fig: trained_model_clustering}, demonstrates the effect of PLMs on NS sampling by LEMON where the entities are well-separated in different clusters(see the left side of Figure~\ref{fig: trained_model_clustering}).
However, there are overlapping entities between the two groups in the case that KGE model are trained by Uniform (see the right side of Figure~\ref{fig: trained_model_clustering}).
% \begin{figure*}[ht!]
% %need a high res picture
%     %\centering
%     \includegraphics[width=1\textwidth]{ECML/clustering _lm.png}
%     %\setlength{\belowcaptionskip}{-10pt}
%     \caption{Clustering visualization from the embedding vectors obtained from Sentence Transformer.}
%     \label{fig: trained_embedding_PLM}
% \end{figure*}
In Figure~\ref{fig: trained_model_clustering}, we observed that, there are certain entities that are not clustered in the expected group for Uniform negative sampling technique. Therefore, for the analysis purpose, we have used UMLs dataset due to its exclusivity. 
'manufactured\_device', 'drug\_delivery\_device', 'clinical\_drug', 'medical\_device' and 'research\_device' are the entities (in green) which are classified with 'language', 'molecular\_sequence', 'spacial\_concept' etc for the trained embeddings from uniform negative sampling (in red). In our approach, both the red and green entities belong to two separate group of clusters. 

For the visualization purpose, we used embeddings from a trained checkpoint of both Uniform and LEMON NS methods. 
TSNE~\cite{van2008visualizing} is employed to reduce the dimension of the trained embedding to 2. 
The axes are denoted as TSNE-1, and TSNE-2 which indicate the two reduced dimensions in Figure~\ref{fig: trained_model_clustering}. 
Due to the space limitation, we restricted the number of entity labels that appear in the visualization to a subset of the entire entity set. 
For clustering the entity embeddings, K-means++ has been trained till 1000 iterations.   
We obtained the KGE embeddings from the RotatE model. 
The hyperparameters used in RotatE this purpose are as follows: $\mathcal{D}$=200, $\mathcal{B}$ = 64, $\mathcal{K}$ = 6, $\mathcal{N}$ = 3, $\gamma $ = 12, $\mathcal{H}$ = 2 and, $\alpha$ = 0.1.

\subsection{Influence of PLMs: Structure vs. Textual Knowledge}% vs. Language Dominance}
\begin{figure*}[ht!]
\centering
\begin{subfigure}{}
    \includegraphics[trim={2.2cm 1cm 2.2cm 1cm},width=\textwidth]{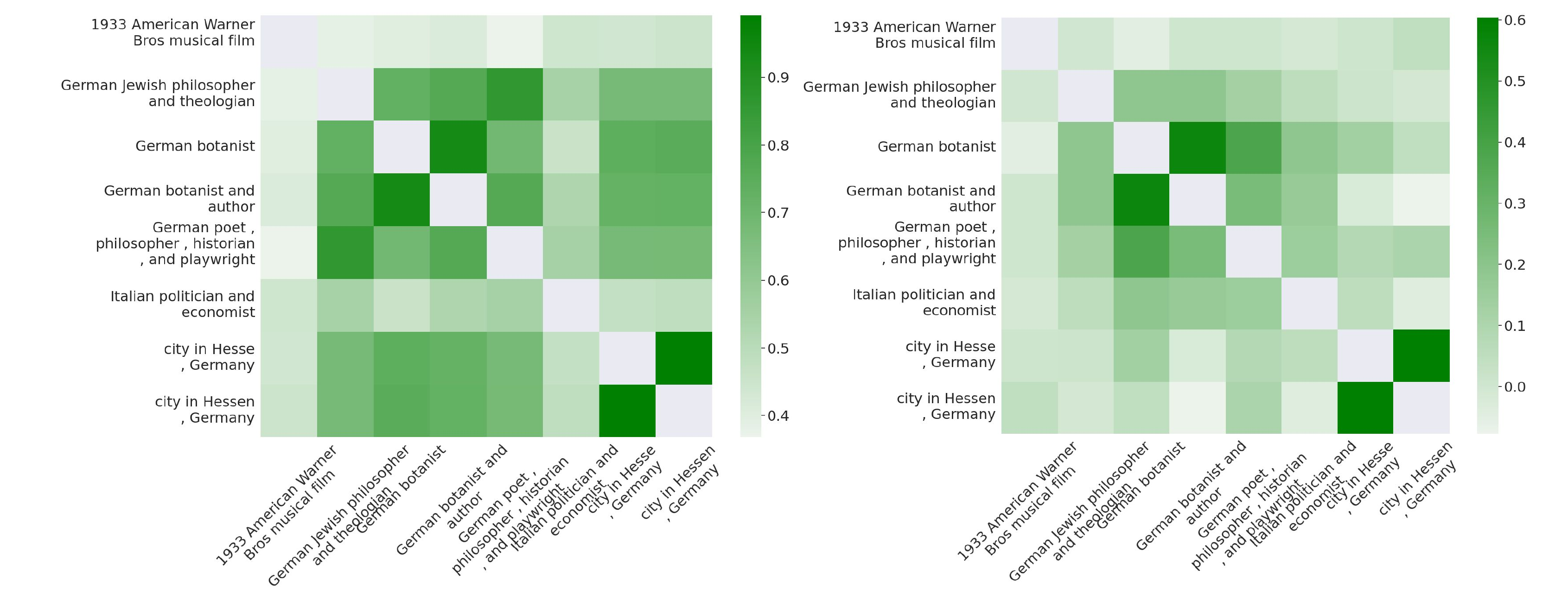}
  \caption{Difference between words from different clusters in CODEx-m dataset.}
  \label{fig: heatmap_comparison_codex}
\end{subfigure}
\begin{subfigure}{}
  \includegraphics[trim={2.2cm 1cm 3cm 0.5cm},width=\textwidth]{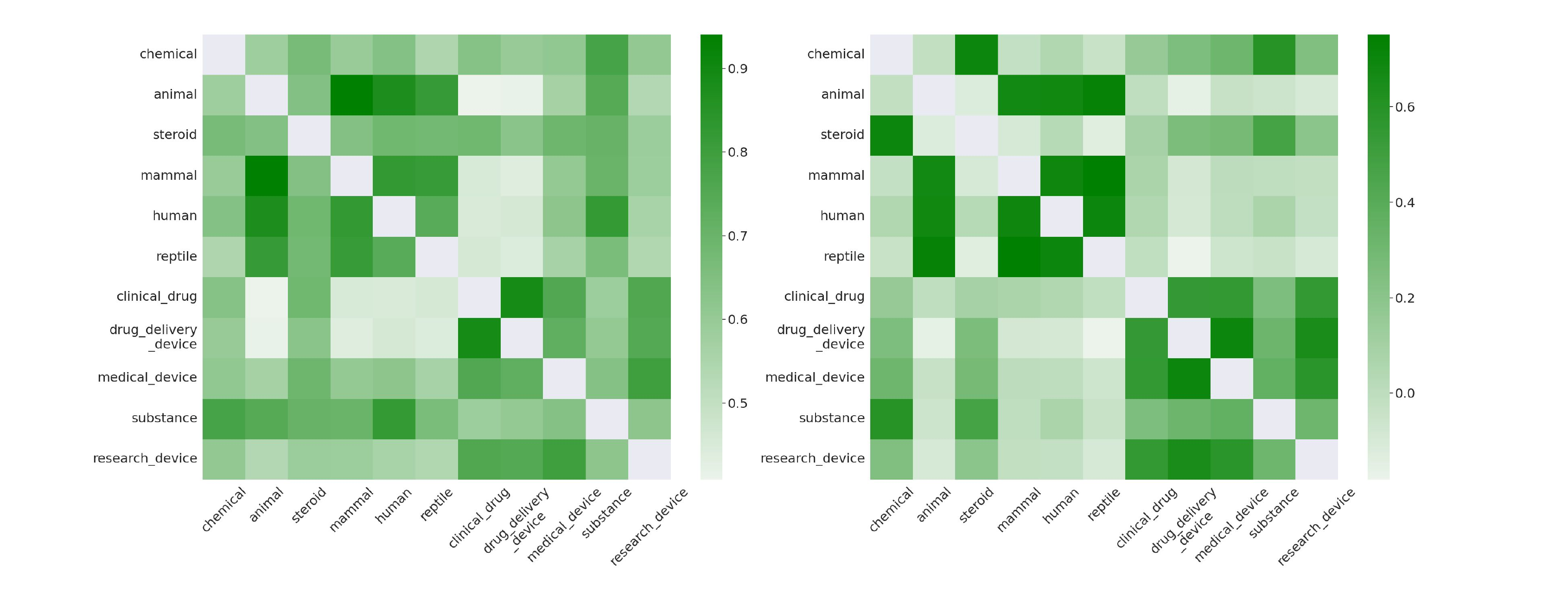}
  \caption{Correlation between words from different clusters in UMLS dataset.}
  \label{fig: heatmap_comparison_umls}
\end{subfigure}
\caption[]{Comparison of KGE embedding (right) vs. Sentence-BERT (left) in both pair of figures. (a) embeddings differs between PLM and KGEs. (b) embeddings of PLMs and KGEs are aligned.}
\end{figure*}

% KGE data consists of varying level of information from text. 
% Often the entity text may not hold enough information to obtain a better representation from the language models. % but when 
During our experiment we observed that, the impact of using PLMs in KGEs depend on the richness of the textual information. 
%On the contrary, the symbolic representation of these entities and relations, which are considered as a graph structure, is more understandable to the KGE models. 
%It is because the structure of the data is more meaningful than the structure of the corresponding text. 
Consider an example from the CoDEx-M dataset, the pre-trained embedding representation of the entity $e$ with text \textit{"1922 1991 country in Europe and Asia"}, carries less meaningful information about the context.
In such cases, the KGE model benefits from its own capability in learning graph structure that preforms better.
%However, the structural representation such as a triple \textit{"(1922 1991 country in Europe and Asia, diplomatic relations of the country, republic in South Asia)"}; helps the KGE model to understand  better the context of the entity \textit{"1922 1991 country in Europe and Asia"}.
We call it \textit{structurally dominant KG} (i.e., CoDEx-M -- see Figure \ref{fig: heatmap_comparison_codex}), since the meaning of the text is not very relevant to the corresponding triple. 
On the other hand, for some KGs (i.e., UMLS) the PLM embeddings not only aligned with the structure-based embeddings, but also provide more informative candidates (Figures \ref{fig: heatmap_comparison_umls}). 
%which are structurally identical in terms of their textual meaning and the structure, can be considered as balanced (i.e., UMLS)
%On the other hand, there are KGs which are structurally identical in terms of their textual meaning and the structure, can be considered as balanced (i.e., UMLS). 
% Regarding above particular assessments, Figures \ref{fig: heatmap_comparison_umls} and \ref{fig: heatmap_comparison_codex} can be considered. 
For instance, the pre-trained embeddings from Sentence-BERT (left) shows similarity with the one from a KGE model (right) (i.e, ``mammal'' and ``human``). 
The heatmap of the following entities also confirm the claim: 'clinical\_drug', 'drug\_delivery\_device', and 'medical\_device' having similarity in heatmap makes the pre-trained entities from the PLMs agree with the one from a trained KGE model. 
Hence, this dataset can be considered as balanced.
%Regarding above particular assessments, 
%More example cases are visualized in Figures \ref{fig: heatmap_comparison_umls} and \ref{fig: heatmap_comparison_codex}. Similarly, in Figure \ref{fig: heatmap_comparison_umls}, the pretrained embedding from Sentence-BERT (left) shows similarity to the one from the trained KGE model (right). Specifically, if we consider the heatmap of mammal and human. Another comparison can be between 'clinical\_drug', 'drug\_delivery\_device', and 'medical\_device'. 
%Having similarity in heatmap makes the pre-trained entities from the PLMs agree with the one which is trained from KGE model. Hence this data can be called as balanced. 

% Regarding above particular assessment, Figures \ref{fig: heatmap_comparison_umls} and \ref{fig: heatmap_comparison_codex} can be considered. 
%In figure \ref{fig: heatmap_comparison_umls}, the pre-trained embedding from BERT (left) shows similarity in heatmap visualization  with the one trained from KGE model (right). 
%Specifically if we consider the heatmap of mammal and human. Another comparison can be between 'clinical\_drug', 'drug\_delivery\_device', and 'medical\_device'. Having similarity in heatmap makes the pretrained entities from the PLMs agree with the one which is trained from KGE model. Hence this data can be called as balanced. 
Figure \ref{fig: heatmap_comparison_codex} demonstrates dissimilarities between the entity embeddings from the PLMs and KGE model.
Consider the following entities: 'German Jewish philosopher and theologian', 'German poet , philosopher , historian , and playwright'.
Though they show high similarity measures in the heatmap of PLMs, but their similarity values are relatively low in the KGE model-based heatmap. 
For instance, the information provided in the PLM could capture the location information very well. 
However, it does not correspond to the information from the structured KG. 
Even though often the embeddings from the PLMs corresponds to the structural information in KGE (i.e, 'city in Hesse, Germany', city in Hessen, Germany), in many cases entities which are similar, the embedding from the PLM end up in a different cluster in the KGE model. 
This happens due to their differences in symbolic structural information. 
In the cases of structural dominance the quality of the clusters reduces and fetches less meaningful negative samples in our case. 
Figure \ref{fig: heatmap_comparison_umls} demonstrates a balanced scenario where the structural and contextual information from the text exists, which we observed from the patterns in the heatmap. 
We noticed several exmples such as 'animal', 'mammal', and 'human' that have higher cosine similarity among themselves in both of the heatmaps. 
% Similar examples are noticeable from the entities 'clinical\_drug', 'drug\_delivery\_device' and medical\_device. Same can be seen in 'substance' and 'chemical'. 
On the other hand less similarity exists in 'mammal' and 'clinical \_drug' in both the images.

\end{document}